\documentclass[11pt]{article}
\usepackage{acl2016}
\usepackage{times}
\usepackage{latexsym}
\usepackage{color}
\usepackage{graphics}
\usepackage{graphicx}

\aclfinalcopy 

\title{Universal Dependencies for Learner English}

\author{Yevgeni Berzak \\ CSAIL MIT \\  berzak@mit.edu \\
        \And Jessica Kenney \\ EECS \& Linguistics MIT \\ jessk@mit.edu \\ 
	\And Carolyn Spadine \\ Linguistics MIT \\ cspadine@mit.edu \\
	\And Jing Xian Wang \\ EECS MIT \\ jxwang@mit.edu \\
	\AND Lucia Lam \\ MECHE MIT \\ lucci@mit.edu \\
	\And Keiko Sophie Mori \\ Linguistics MIT \\ ksmori@mit.edu \\
	\And Sebastian Garza \\ Linguistics MIT \\ sjgarza@mit.edu \\
	\And Boris Katz \\ CSAIL MIT \\ boris@mit.edu}

\date{}

\begin{document}

\maketitle

\begin{abstract}
We introduce the Treebank of Learner English (TLE), 
the first publicly available syntactic treebank for 
English as a Second Language (ESL). 
The TLE provides manually annotated POS tags and Universal Dependency (UD) 
trees for 5,124 sentences from the Cambridge First Certificate 
in English (FCE) corpus. The UD annotations are tied to 
a pre-existing error annotation of the FCE, whereby full 
syntactic analyses are provided for both the original and 
error corrected versions of each sentence. Further on, we 
delineate ESL annotation guidelines that allow for consistent 
syntactic treatment of ungrammatical English. Finally, we 
benchmark POS tagging and dependency parsing performance on 
the TLE dataset and measure the effect of grammatical errors 
on parsing accuracy. We envision the treebank to support a 
wide range of linguistic and computational research on second 
language acquisition as well as automatic processing of 
ungrammatical language\footnote{The treebank is available at  
universaldependencies.org. The annotation manual used in this project
and a graphical query engine are available at esltreebank.org.}.
\end{abstract}

\section{Introduction}
\label{sect:introduction}

The majority of the English text available worldwide is generated by non-native 
speakers \cite{crystal2003English}. Such texts introduce a variety of challenges, 
most notably grammatical errors, and are of paramount importance for the 
scientific study of language acquisition as well as for NLP. Despite the ubiquity 
of non-native English, there is currently no publicly available syntactic treebank 
for English as a Second Language (ESL). 

To address this shortcoming, we present the Treebank of Learner English (TLE), 
a first of its kind resource for non-native English, containing 5,124 sentences manually 
annotated with POS tags and dependency trees. The TLE sentences are drawn from 
the FCE dataset \cite{fcecorpus2011}, and authored by English learners 
from 10 different native language backgrounds. The treebank uses the Universal 
Dependencies (UD) formalism \cite{de2014universal,ud2016}, which provides 
a unified annotation framework across different languages and is geared towards 
multilingual NLP \cite{mcdonald2013universal}. This characteristic allows our 
treebank to support computational analysis of ESL using not only English based 
but also multilingual approaches which seek to relate ESL phenomena to native 
language syntax.

While the annotation inventory and guidelines are defined by the English UD formalism, 
we build on previous work in learner language analysis \cite{diaz2010towards,ragheb2013} 
to formulate an additional set of annotation conventions aiming at a uniform 
treatment of ungrammatical learner language. Our annotation scheme uses a two-layer analysis, 
whereby a distinct syntactic annotation is provided for the \emph{original} and the 
\emph{corrected} version of each sentence. This approach is enabled by a pre-existing 
error annotation of the FCE \cite{nicholls2003clc} which is used to generate an error 
corrected variant of the dataset. Our inter-annotator agreement results 
provide evidence for the ability of the annotation scheme
to support consistent annotation of ungrammatical structures.

Finally, a corpus that is annotated with both grammatical errors and syntactic dependencies
paves the way for empirical investigation of the relation between 
grammaticality and syntax. Understanding this relation is vital for 
improving tagging and parsing performance on learner language \cite{geertzen2013},
syntax based grammatical error correction \cite{tetreault2010using,ng2014conll}, and
many other fundamental challenges in NLP. In this work, we take the first step 
in this direction by benchmarking tagging and parsing accuracy on our dataset under different 
training regimes, and obtaining several estimates for the impact of grammatical errors on 
these tasks.

To summarize, this paper presents three contributions. First, we introduce the first
large scale syntactic treebank for ESL, manually annotated with POS tags and universal 
dependencies. Second, we describe a linguistically motivated annotation scheme for 
ungrammatical learner English and provide empirical support for its consistency via inter-annotator 
agreement analysis. Third, we benchmark a state of the art parser on our dataset 
and estimate the influence of grammatical errors on the accuracy of automatic POS 
tagging and dependency parsing.

The remainder of this paper is structured as follows. We start by presenting an 
overview of the treebank in section \ref{treebank_overview}. In sections 
\ref{annotator_training} and \ref{annotation_procedure} we provide background 
information on the annotation project, and review the main annotation stages 
leading to the current form of the dataset. The ESL annotation guidelines 
are summarized in section \ref{esl_guidelines}. 
Inter-annotator agreement analysis is presented in section \ref{agreement}, followed by parsing
experiments in section \ref{parsing_benchmarks}. Finally, we review related work 
in section \ref{related_work} and present the conclusion in section \ref{conclusion}.

\section{Treebank Overview}
\label{treebank_overview}

The TLE currently contains 5,124 sentences (97,681 tokens) with POS tag and dependency 
annotations in the English Universal Dependencies (UD) formalism \cite{de2014universal,ud2016}. 
The sentences were obtained from the FCE corpus \cite{fcecorpus2011}, a 
collection of upper intermediate English learner essays, containing error 
annotations with 75 error categories \cite{nicholls2003clc}. Sentence level 
segmentation was performed using an adaptation of the NLTK sentence 
tokenizer\footnote{http://www.nltk.org/api/nltk.tokenize.html}. Under-segmented 
sentences were split further manually. Word level tokenization was generated 
using the Stanford PTB word tokenizer\footnote{http://nlp.stanford.edu/software/tokenizer.shtml}.

The treebank represents learners with 10 different native language backgrounds: 
Chinese, French, German, Italian, Japanese, Korean, Portuguese, Spanish, Russian and 
Turkish. For every native language, we randomly sampled 500 automatically segmented 
sentences, under the constraint that selected sentences have to contain at least one 
grammatical error that is not punctuation or spelling.

The TLE annotations are provided in two versions. 
The first version is the  \emph{original sentence} authored by the learner, containing 
grammatical errors. The second, \emph{corrected sentence} version, is a grammatical 
variant of the original sentence, generated by correcting all the grammatical errors 
in the sentence according to the manual error annotation provided in the FCE dataset.
The resulting corrected sentences constitute a parallel corpus of standard English. 
Table \ref{corpus_stats} presents basic statistics of both versions of the annotated sentences.

\begin{table}[!ht]
\resizebox{\columnwidth}{!}{%
\begin{tabular}{lll}
  & original & corrected \\
  \hline
  sentences                       & 5,124             & 5,124 \\
  tokens                          & 97,681           & 98,976 \\
  sentence length                 & 19.06 (std 9.47) & 19.32 (std 9.59) \\
  errors per sentence         & 2.67 (std 1.9)  & - \\ 
   \cline{2-3}
  authors                   &  \multicolumn{2}{c}{924} \\ 
  native languages          &  \multicolumn{2}{c}{10} \\  
 \hline
\end{tabular}
}
\caption{Statistics of the TLE. Standard deviations are denoted in parenthesis.}
\label{corpus_stats}
\end{table}
 
To avoid potential annotation biases, the annotations 
of the treebank were created manually \emph{from scratch}, without 
utilizing any automatic annotation tools. To further assure annotation quality, 
each annotated sentence was reviewed by two additional annotators. To the best 
of our knowledge, TLE is the first large scale English treebank constructed in a 
completely manual fashion.

\section{Annotator Training}
\label{annotator_training}

The treebank was annotated by six students, five undergraduates 
and one graduate. Among the undergraduates, three are linguistics 
majors and two are engineering majors with a linguistic minor. 
The graduate student is a linguist specializing in syntax. An 
additional graduate student in NLP participated in the final debugging 
of the dataset.

Prior to annotating the treebank sentences, the annotators were trained for about
8 weeks. During the training, the annotators attended 
tutorials on dependency grammars, and learned the English UD 
guidelines\footnote{http://universaldependencies.org/\#en}, 
the Penn Treebank POS guidelines \cite{ptbpos}, the grammatical error annotation 
scheme of the FCE \cite{nicholls2003clc}, as well as the ESL 
guidelines described in section \ref{esl_guidelines} and in the annotation manual. 

Furthermore, the annotators completed six annotation exercises, in which they 
were required to annotate POS tags and dependencies for practice sentences from scratch. 
The exercises were done individually, and were followed by group meetings in which 
annotation disagreements were discussed and resolved. Each of the first three 
exercises consisted of 20 sentences from the UD gold 
standard for English, the English Web Treebank (EWT) \cite{silveira2014gold}. 
The remaining three exercises contained 20-30 ESL sentences from the 
FCE. Many of the ESL guidelines were introduced or refined based on the 
disagreements in the ESL practice exercises and the subsequent group discussions. 
Several additional guidelines were introduced in the course of the annotation 
process.

During the training period, the annotators also learned to use a search tool that 
enables formulating queries over word and POS tag sequences as regular expressions 
and obtaining their annotation statistics in the EWT. 
After experimenting with both textual and graphical interfaces for performing the 
annotations, we converged on a simple text based format described in section 
\ref{subsec:procedure-annotation}, where the annotations were filled in using a 
spreadsheet or a text editor, and tested with a script for detecting annotation 
typos. The annotators continued to meet and discuss annotation issues
on a weekly basis throughout the entire duration of the project.

\section{Annotation Procedure}
\label{annotation_procedure}
The formation of the treebank was carried out in four steps: annotation, review, 
disagreement resolution and targeted debugging.

\subsection{Annotation} 
\label{subsec:procedure-annotation}

In the first stage, the annotators were given sentences for annotation from scratch. 
We use a CoNLL based textual template in which each word is 
annotated in a separate line. Each line contains 6 columns, the first of 
which has the word index (IND) and the second the word itself (WORD). 
The remaining four columns had to be filled in with a Universal POS 
tag (UPOS), a Penn Treebank POS tag (POS), a head word index (HIND) and a dependency 
relation (REL) according to version 1 of the English UD guidelines. 

The annotation section of the sentence is preceded by a metadata header.
The first field in this header, denoted with SENT, contains the FCE error coded version of 
the sentence. The annotators were instructed to verify the error annotation, 
and add new error annotations if needed.  
Corrections to the sentence segmentation are specified in the SEGMENT 
field\footnote{The released version of the treebank splits the sentences 
according to the markings in the SEGMENT field when those apply both
to the original and corrected versions of the sentence. Resulting segments 
without grammatical errors in the original version are currently discarded.}.
Further down, the field TYPO is designated for literal annotation of spelling 
errors and ill formed words that happen to form valid words (see section \ref{subsec:exceptions}). 

The example below presents a pre-annotated original sentence given to an annotator.

\begin{scriptsize}
\texttt{
\\
\#SENT=That time I had to sleep in <ns type= "MD"><c>a</c></ns> tent. \\
\#SEGMENT= \\
\#TYPO=
\begin{tabbing}
\hspace*{1cm}\=\hspace*{1.1cm}\=\hspace*{1.1cm}\=\hspace*{1.1cm}\=\hspace*{1.1cm}\= \kill
\#IND    \>WORD    \>UPOS    \>POS     \>HIND    \>REL \\
1       \>That    \>         \>        \>        \> \\
2       \>time    \>         \>       \>         \> \\
3       \>I       \>    \>      \>       \> \\
4       \>had     \>    \>     \>       \> \\ 
5       \>to      \>    \>      \>       \> \\
6       \>sleep   \>    \>      \>       \> \\
7       \>in      \>   \>      \>       \> \\
8       \>tent    \>    \>      \>       \> \\
9       \>.       \>   \>   \>       \>
\end{tabbing}
}
\end{scriptsize}

Upon completion of the original sentence, the annotators proceeded 
to annotate the corrected sentence version. To reduce annotation time, 
annotators used a script that copies over annotations from the original sentence 
and updates head indices of tokens that appear in both sentence versions. 
Head indices and relation labels were filled in only if the head word 
of the token appeared in both the original and corrected sentence versions.
Tokens with automatically filled annotations included an additional 
\# sign in a seventh column of each word's annotation. The \# signs 
had to be removed, and the corresponding annotations either approved 
or changed as appropriate. Tokens that did not appear in the original 
sentence version were annotated from scratch. 
 
\subsection{Review}
 
All annotated sentences were randomly assigned to a second 
annotator (henceforth \emph{reviewer}), in a double blind manner. 
The reviewer's task was to mark all the annotations that they would 
have annotated differently. To assist the review process, we 
compiled a list of common annotation errors, available in the 
released annotation manual. 

The annotations were reviewed using an \emph{active} 
editing scheme in which an explicit action was required for all the 
existing annotations. The scheme was introduced to prevent reviewers
from overlooking annotation issues due to passive approval. 
Specifically, an additional \# sign was added at the seventh column of each 
token's annotation. The reviewer then had to either ``sign off'' on the 
existing annotation by erasing the \# sign, or provide an alternative 
annotation following the \# sign.

\subsection{Disagreement Resolution}
\label{subsec:resolution}

In the final stage of the annotation process all annotator-reviewer 
disagreements were resolved by a third annotator (henceforth \emph{judge}), 
whose main task was to decide in favor of the annotator or the reviewer. 
Similarly to the review process, the judging task was carried out in a 
double blind manner. Judges were allowed to resolve annotator-reviewer 
disagreements with a third alternative, as well as introduce new corrections 
for annotation issues overlooked by the reviewers. 

Another task performed by the judges was to mark acceptable \emph{alternative 
annotations} for ambiguous structures determined through review disagreements 
or otherwise present in the sentence. These annotations were specified in an additional
metadata field called AMBIGUITY. The ambiguity markings are provided along with the 
resolved version of the annotations. 

\subsection{Final Debugging}

After applying the resolutions produced by the judges, we queried the corpus with
debugging tests for specific linguistics constructions. 
This additional testing phase further reduced the number of annotation 
errors and inconsistencies in the treebank. Including the training period, 
the treebank creation lasted over a year, with an aggregate of more than 2,000 
annotation hours.

\section{Annotation Scheme for ESL}
\label{esl_guidelines}

Our annotations use the existing inventory of English UD POS tags and dependency 
relations, and follow the standard UD annotation guidelines for English. 
However, these guidelines were formulated with grammatical usage of English 
in mind and do not cover non canonical syntactic structures 
arising due to grammatical errors\footnote{The English UD guidelines do address 
several issues encountered in informal genres, such as the relation ``goeswith'', 
which is used for fragmented words resulting from typos.}. 
To encourage consistent and linguistically motivated annotation 
of such structures, we formulated a complementary set of ESL annotation guidelines. 

Our ESL annotation guidelines follow the general principle of 
\emph{literal reading}, which emphasizes syntactic analysis according to 
the observed language usage. This strategy continues a line of work 
in SLA which advocates for centering analysis of learner language 
around morpho-syntactic surface evidence \cite{ragheb2012defining,ragheb2013}. 
Similarly to our framework, which includes a parallel annotation of corrected sentences, 
such strategies are often presented in the context of multi-layer annotation 
schemes that also account for error corrected sentence forms \cite{hirschmann2007syntactic,diaz2010towards,rosen2014evaluating}.

Deploying a strategy of literal annotation within UD, a formalism which 
enforces cross-linguistic consistency of annotations, will enable meaningful 
comparisons between non-canonical structures in English and canonical structures in the 
author's native language. As a result, a key novel characteristic of our treebank is 
its ability to support cross-lingual studies of learner language. 

\subsection{Literal Annotation}

With respect to POS tagging, literal annotation implies adhering as much as 
possible to the observed morphological forms of the words. Syntactically, argument 
structure is annotated according to the usage of the word rather than its typical 
distribution in the relevant context. The following list of conventions defines 
the notion of literal reading for some of the common non canonical structures 
associated with grammatical errors. 

\subsubsection*{Argument Structure}

\emph{Extraneous prepositions} 
We annotate all nominal dependents introduced by extraneous 
prepositions as nominal modifiers. In the following sentence, 
``him'' is marked as a nominal modifier (\emph{nmod}) 
instead of an indirect object (\emph{iobj}) of ``give''.

\begin{scriptsize}
\texttt{
\\
\#SENT=...I had to give <ns type="UT"><i>to</i> </ns> him water...
\begin{tabbing}
	\hspace*{1.1cm}\=\hspace*{1.5cm}\=\hspace*{1.3cm}\=\hspace*{1.1cm}\=\hspace*{1.1cm}\= \kill
	...	\>	\>	\>	\>	\>    \\
	21	\>I	\>PRON	\>PRP	\>22	\>nsubj\\
	22	\>had \>VERB	\>VBD	\>5	\>parataxis\\
	23	\>to	\>PART	\>TO	\>24	\>mark\\
	24	\>\bf give	\>VERB	\>VB	\>22	\>xcomp\\
	25	\>to	\>ADP	\>IN	 \>26	\>case\\
	26    	\>\bf him\>PRON	\>PRP	\>24	\>\bf nmod \\
	27	\>water	\>NOUN	\>NN	\>24	\>dobj\\
	...	\>	\>	\>	\>	\>    
	\end{tabbing}
	}
\end{scriptsize}

\emph{Omitted prepositions} We treat nominal dependents 
of a predicate that are lacking a preposition as arguments rather than nominal modifiers.
In the example below, ``money'' is marked as a direct object (\emph{dobj}) instead 
of a nominal modifier (\emph{nmod}) of ``ask''. As ``you'' functions in this context 
as a second argument of ``ask'', it is annotated as an indirect object (\emph{iobj}) 
instead of a direct object (\emph{dobj}).

\begin{scriptsize}
\texttt{
\\
\#SENT=...I have to ask you <ns type="MT"> <c>for</c></ns> the money <ns type= "RT"> <i>of</i><c>for</c></ns> the tickets back.
	\begin{tabbing}
  	\hspace*{1.1cm}\=\hspace*{1.5cm}\=\hspace*{1.3cm}\=\hspace*{1.1cm}\=\hspace*{1.1cm}\= \kill
	...	\>	\>	\>	\>	\>    \\
	12	\>I	\>PRON	\>PRP	\>13	\>nsubj\\
	13	\>have	\>VERB	\>VBP	\>2	\>conj\\
	14	\>to	\>PART	\>TO	\>15	\>mark\\
	15	\>\bf ask	\>VERB	\>VB	\>13	\>xcomp\\
	16	\>\bf you	\>PRON	\>PRP	\>15	\>\bf iobj \\ 
	17	\>the	\>DET	\>DT	\>18	\>det \\
	18	\>\bf money	\>NOUN	\>NN	\>15	\>\bf dobj \\ 
	19	\>of	\>ADP	\>IN	\>21	\>case \\
	20	\>the	\>DET	\>DT	\>21	\>det \\
	21	\>tickets\>NOUN	\>NNS	\>18	\>nmod \\
	22	\>back	\>ADV	\>RB	\>15	\>advmod \\
	23	\>.	\>PUNCT	\>.	\>2	\>punct
	\end{tabbing}
	}
\end{scriptsize}

\subsubsection*{Tense}
Cases of erroneous tense usage are annotated according to the morphological tense 
of the verb. For example, below we annotate ``shopping'' with present participle VBG, 
while the correction ``shop'' is annotated in the corrected version of the sentence as VBP.

\begin{scriptsize}
\texttt{
\\
\#SENT=...when you  <ns type="TV"><i>shopping</i> <c>shop</c></ns>...  
\begin{tabbing}
\hspace*{1.1cm}\=\hspace*{1.5cm}\=\hspace*{1.3cm}\=\hspace*{1.1cm}\=\hspace*{1.1cm}\= \kill
...	\>	\>	\>	\>	\>    \\
4       \>when       \>ADV    \>WRB     \>6       \>advmod \\
5       \>you      \>PRON     \>PRP      \>6       \>nsubj \\
6       \>\bf shopping  \>VERB     \>\bf VBG      \>12       \>advcl\\
...	\>	\>	\>	\>	\>    
\end{tabbing}
} \end{scriptsize}

\subsubsection*{Word Formation}

Erroneous word formations that are contextually plausible and can be assigned with a
PTB tag are annotated literally. In the following example, ``stuffs'' is
handled as a plural count noun.

\begin{scriptsize}
\texttt{
\\
\#SENT=...into fashionable <ns type="CN"> <i>stuffs</i><c>stuff</c></ns>...
	\begin{tabbing}
  	\hspace*{1.1cm}\=\hspace*{1.8cm}\=\hspace*{1.3cm}\=\hspace*{1.1cm}\=\hspace*{1.1cm}\= \kill
...	\>	\>	\>	\>	\> \\
7	\>into	\>ADP	\>IN	\>9	\>case \\
8	\>fashionable	\>ADJ	\>JJ	\>9	\>amod \\
9	\>\bf stuffs	\>NOUN	\>\bf NNS	\>2	\>ccomp \\
...	\>	\>	\>	\>	\>
	\end{tabbing}
	}
\end{scriptsize}

Similarly, in the example below we annotate ``necessaryiest'' as a superlative.

\begin{scriptsize}
\texttt{
\\
\#SENT=The necessaryiest things...
	\begin{tabbing}
  	\hspace*{1.1cm}\=\hspace*{2.1cm}\=\hspace*{1.3cm}\=\hspace*{1.1cm}\=\hspace*{1.1cm}\= \kill
1	\>The	\>DET	\>DT	\>3	\>det \\
2	\> \bf necessaryiest	\>ADJ	\> \bf JJS	\>3	\>amod \\
3	\>things	\>NOUN	\>NNS	\>0	\>root \\
...	\>	\>	\>	\>	\>
	\end{tabbing}
	}
\end{scriptsize}

\subsection{Exceptions to Literal Annotation}
\label{subsec:exceptions}

Although our general annotation strategy for ESL follows literal sentence readings,
several types of word formation errors make such readings uninformative or impossible, 
essentially forcing certain words to be annotated using some degree of interpretation 
\cite{rosen2010syntactic}. We hence annotate the following cases in the original sentence 
according to an interpretation of an intended word meaning, obtained from the FCE error 
correction.

\subsubsection*{Spelling} 
Spelling errors are annotated according to the correctly spelled 
version of the word. To support error analysis of automatic annotation tools, 
misspelled words that happen to form valid words are annotated in the metadata field TYPO 
for POS tags with respect to the most common usage of the misspelled word form.
In the example below, the TYPO field contains the typical POS annotation of 
``where'', which is clearly unintended in the context of the sentence. 

\begin{scriptsize}
\texttt{
\\
\#SENT=...we <ns type="SX"><i>where</i> <c>were</c></ns> invited to visit...\\
\textbf{\#TYPO=5 ADV WRB}
\begin{tabbing}
\hspace*{1.1cm}\=\hspace*{1.5cm}\=\hspace*{1.3cm}\=\hspace*{1.1cm}\=\hspace*{1.1cm}\= \kill
...      \>   \>     \>      \>       \> \\
4       \>we      \>PRON    \>PRP     \>6       \>nsubjpass \\
5       \>\bf where   \>\bf AUX     \>\bf VBD     \>\bf 6       \>\bf auxpass \\
6       \>invited \>VERB    \>VBN     \>0       \>root \\
7       \>to      \>PART    \>TO      \>8       \>mark \\
8       \>visit   \>VERB    \>VB      \>6       \>xcomp \\
...	\>	\>	\>	\>	\>
\end{tabbing}
}
\end{scriptsize}

\subsubsection*{Word Formation}

Erroneous word formations that cannot be assigned with an
existing PTB tag are annotated with respect to the correct word form.

\begin{scriptsize}
\texttt{
\\
\#SENT=I am  <ns type="IV"><i>writting</i> <c>writing</c></ns>...
\begin{tabbing}
\hspace*{1.1cm}\=\hspace*{1.5cm}\=\hspace*{1.3cm}\=\hspace*{1.1cm}\=\hspace*{1.1cm}\= \kill
1	\>I	\>PRON	\>PRP	\>3	\>nsubj \\
2	\>am	\>AUX	\>VBP	\>3	\>aux \\
3	\> \bf writting	\>VERB	\>VBG	\>0	\>root\\
...	\>	\>	\>	\>	\>
\end{tabbing}
}
\end{scriptsize}

In particular, ill formed adjectives that have a plural suffix receive a standard 
adjectival POS tag. When applicable, such cases also receive an additional marking for unnecessary agreement 
in the error annotation using the attribute ``ua''.

\begin{scriptsize}
\texttt{
\\
\#SENT=...<ns type="IJ" \textbf{ua=true}> <i>interestings</i><c>interesting</c></ns> things...
\begin{tabbing}
\hspace*{1.1cm}\=\hspace*{1.5cm}\=\hspace*{1.3cm}\=\hspace*{1.1cm}\=\hspace*{1.1cm}\= \kill
...	\>	\>	\>	\>	\> \\
6	\>\bf interestings	\>ADJ	\>\bf JJ	\>7	\>amod \\
7	\>things	\>NOUN	\>NNS	\>3	dobj \\
...	\>	\>	\>	\>	\>
\end{tabbing}
}
\end{scriptsize}

Wrong word formations that result in a valid, but contextually implausible word form 
are also annotated according to the word correction. In the example below, the nominal 
form ``sale'' is likely to be an unintended result of an ill formed verb. Similarly to 
spelling errors that result in valid words, we mark the typical literal POS annotation 
in the TYPO metadata field.

\begin{scriptsize}
\texttt{
\\
\#SENT=...they do not <ns type="DV"><i>sale</i> <c>sell</c></ns> them...\\
\textbf{\#TYPO=15 NOUN NN}
\begin{tabbing}
\hspace*{1.1cm}\=\hspace*{1.5cm}\=\hspace*{1.3cm}\=\hspace*{1.1cm}\=\hspace*{1.1cm}\= \kill
...   \>      \>           \>         \>         \> \\
12      \>they    \>PRON     \>PRP    \>15      \>nsubj \\
13      \>do      \>AUX     \>VBP     \>15      \>aux \\
14      \>not     \>PART      \>RB    \>15      \>neg \\
15      \>\bf sale    \> \bf VERB      \>\bf VB    \>\bf 0       \> \bf root \\
16      \>them    \>PRON     \>PRP    \>15      \>dobj \\
...	\>	\>	\>	\>	\>   
\end{tabbing}
}
\end{scriptsize}

Taken together, our ESL conventions cover many of the annotation challenges related 
to grammatical errors present in the TLE. In addition to the presented overview, the 
complete manual of ESL guidelines used by the annotators is 
publicly available. The manual contains further details 
on our annotation scheme, additional annotation guidelines and a list of 
common annotation errors. We plan to extend and refine these guidelines in 
future releases of the treebank.

\section{Editing Agreement}
\label{agreement}

We utilize our two step review process to estimate agreement
rates between annotators\footnote{All experimental results on agreement 
and parsing exclude punctuation tokens.}. We measure agreement as the fraction
of annotation tokens approved by the editor. Table \ref{agreement_table} 
presents the agreement between annotators and reviewers, as well as the agreement between
reviewers and the judges. 
Agreement measurements are provided for both the original the corrected versions 
of the dataset.

\begin{table}[ht!]
\resizebox{\columnwidth}{!}{%
\begin{tabular}{lllll} \hline
\bf Annotator-Reviewer  & UPOS & POS & HIND & REL  \\ \hline
original	& 98.83 &        98.35 &	97.74&	96.98  \\
corrected 	& 99.02 &	98.61 &	97.97&	97.20  \\ \hline
\bf Reviewer-Judge  &   &  & &  \\ \hline
original	& 99.72 &	99.68 &	99.37 &	99.15 \\
corrected	& 99.80 &	99.77 &	99.45 &	99.28 \\ \hline
\end{tabular}%
}
\caption{Inter-annotator agreement on the entire TLE corpus. Agreement is measured as 
the fraction of tokens that remain unchanged after an editing round. 
The four evaluation columns correspond to universal POS tags, PTB POS tags, 
unlabeled attachment, and dependency labels. Cohen's Kappa scores \cite{kappa} 
for POS tags and dependency labels in all evaluation conditions are above 0.96.}
\label{agreement_table}
\end{table}

Overall, the results indicate a high agreement rate in the two editing tasks. 
Importantly, the gap between the agreement on the original and corrected sentences is small. 
Note that this result is obtained despite the introduction of several ESL annotation 
guidelines in the course of the annotation process, which inevitably increased the number 
of edits related to grammatical errors. We interpret this outcome as evidence 
for the effectiveness of the ESL annotation scheme in supporting consistent annotations 
of learner language.
 
\section{Parsing Experiments}
\label{parsing_benchmarks}

The TLE enables studying parsing for learner language and exploring
relationships between grammatical errors and parsing performance.  
Here, we present parsing benchmarks on our dataset, and provide 
several estimates for the extent to which grammatical errors degrade 
the quality of automatic POS tagging and dependency parsing.

Our first experiment measures tagging and parsing accuracy on the TLE
and approximates the global impact of grammatical errors on automatic annotation  
via performance comparison between the original and error corrected sentence versions. 
In this, and subsequent experiments, we utilize version 2.2 of the Turbo 
tagger and Turbo parser \cite{martins2013}, state of the art tools for statistical 
POS tagging and dependency parsing.  

Table \ref{parsing_results_table} presents tagging and parsing results on a test set of 500 
TLE sentences (9,591 original tokens, 9,700 corrected tokens).
Results are provided for three different training regimes. The first regime uses the 
training portion of version 1.3 of the EWT, the UD English treebank, containing 
12,543 sentences (204,586 tokens). The second training mode uses 4,124 training sentences 
(78,541 original tokens, 79,581 corrected tokens) from the TLE corpus. In the third 
setup we combine these two training corpora. The remaining 500 TLE sentences 
(9,549 original tokens, 9,695 corrected tokens) are allocated to a development set, 
not used in this experiment. Parsing of the test sentences was performed 
on predicted POS tags. 

\begin{table}[ht!]
\resizebox{\columnwidth}{!}{%
\begin{tabular}{lllllllll}
\bf Test set & \bf Train Set &  \bf UPOS & \bf POS & \bf UAS &  \bf LA & \bf LAS \\ \hline
TLE$_{orig}$ & EWT  		& 91.87&  94.28&  86.51&  88.07&  81.44 \\
TLE$_{corr}$ & EWT 		& 92.9&   95.17&  88.37&  89.74&  83.8 \\ \hline
TLE$_{orig}$ & TLE$_{orig}$ 	& 95.88&  94.94&  87.71&  89.26&  83.4 \\
TLE$_{corr}$ & TLE$_{corr}$ 	& 96.92&  95.17&  89.69&  90.92&  85.64 \\ \hline
TLE$_{orig}$ & EWT+TLE$_{orig}$ & 93.33&  95.77&  90.3&   91.09&  86.27 \\ 
TLE$_{corr}$ & EWT+TLE$_{corr}$ & 94.27&  96.48&  92.15&  92.54&  88.3 \\
\end{tabular}%
}
\caption{Tagging and parsing results on a test set of 500 sentences from the TLE corpus. 
EWT is the English UD treebank.
TLE$_{orig}$ are original sentences from the TLE.
TLE$_{corr}$ are the corresponding error corrected sentences.}
\label{parsing_results_table}
\end{table}

The EWT training regime, which uses out of domain texts written in standard English, 
provides the lowest performance on all the evaluation metrics. An additional factor which negatively
affects performance in this regime are systematic differences in the 
EWT annotation of possessive pronouns, expletives and names compared to the UD guidelines,
which are utilized in the TLE. In particular, the EWT annotates possessive pronoun UPOS
as PRON rather than DET, which leads the UPOS results in this setup to be lower than the PTB
POS results. Improved results are obtained using the TLE training data, which, despite 
its smaller size, is closer in genre and syntactic characteristics to the TLE test set. 
The strongest PTB POS tagging and parsing results are obtained by combining the EWT with 
the TLE training data, yielding 95.77 POS accuracy and a UAS of 90.3 on the 
original version of the TLE test set.

The dual annotation of sentences in their original and error corrected forms enables 
estimating the impact of grammatical errors on tagging and parsing by  
examining the performance gaps between the two sentence versions. Averaged across the 
three training conditions, the POS tagging accuracy on the original sentences is lower
than the accuracy on the sentence corrections by 1.0 UPOS and 0.61 POS. 
Parsing performance degrades by 1.9 UAS, 1.59 LA and 2.21 LAS.

To further elucidate the influence of grammatical errors on parsing quality, 
table \ref{parsing_errors_table} compares performance on tokens in the original sentences 
appearing inside grammatical error tags to those appearing outside such tags. 
Although grammatical errors may lead to tagging and parsing errors with respect to 
any element in the sentence, we expect erroneous tokens to be more challenging to analyze
compared to grammatical tokens. 

\begin{table}[ht!]
\resizebox{\columnwidth}{!}{%
\begin{tabular}{lllllllll}
\bf Tokens & \bf Train Set &  \bf UPOS & \bf POS & \bf UAS & \bf LA & \bf LAS \\ \hline
Ungrammatical    	 & EWT 		& 87.97&  88.61&  82.66&  82.66&  74.93\\
Grammatical    		 & EWT 		& 92.62&  95.37&  87.26&  89.11&  82.7 \\ \hline
Ungrammatical   & TLE$_{orig}$ 	& 90.76&  88.68&  83.81&  83.31&  77.22 \\ 
Grammatical 	& TLE$_{orig}$ 	& 96.86&  96.14&  88.46&  90.41&  84.59 \\ \hline
Ungrammatical & EWT+TLE$_{orig}$  	& 89.76&  90.97&  86.32&  85.96&  80.37 \\  
Grammatical & EWT+TLE$_{orig}$  	& 94.02&  96.7&   91.07&  92.08&  87.41 \\  \hline
\end{tabular}%
}
\caption{Tagging and parsing results on the original version of the TLE test set for 
tokens marked with grammatical errors (Ungrammatical) and tokens not marked for 
errors (Grammatical).}
\label{parsing_errors_table}
\end{table}

This comparison indeed reveals 
a substantial difference between the two types of tokens, with an average gap of
5.0 UPOS, 6.65 POS, 4.67 UAS, 6.56 LA and 7.39 LAS.
Note that differently from the global measurements in the first experiment, this analysis, 
which focuses on the local impact of remove/replace errors, suggests a stronger 
effect of grammatical errors on the dependency labels than on the dependency 
structure.

Finally, we measure tagging and parsing performance relative to the fraction 
of sentence tokens marked with grammatical errors. Similarly to the previous experiment,
this analysis focuses on remove/replace rather than insert errors. 

\begin{figure}[h]  
\label{parsing_curves}
  \centering
    \includegraphics[width=0.48\textwidth]{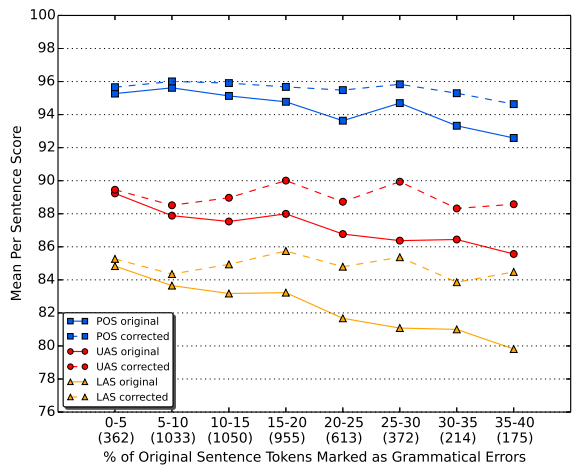}
\caption{Mean per sentence POS accuracy, UAS and LAS of the Turbo tagger and Turbo parser, 
as a function of the percentage of original sentence tokens marked with grammatical errors. 
The tagger and the parser are trained on the EWT corpus, and tested on all 5,124 sentences of the TLE. 
Points connected by continuous lines denote performance on the original TLE sentences. 
Points connected by dashed lines denote performance on the corresponding error corrected sentences. 
The number of sentences whose errors fall within each percentage range appears in parenthesis.}
\end{figure}

Figure 1 presents the average sentential performance as a function of the percentage of 
tokens in the original sentence marked with grammatical errors.   
In this experiment, we train the parser on the EWT training set and test on the entire TLE corpus.
Performance curves are presented for POS, UAS and LAS on the original and error corrected
versions of the annotations. We observe that while the performance on the corrected sentences is close
to constant, original sentence performance is decreasing as the 
percentage of the erroneous tokens in the sentence grows.

Overall, our results suggest a negative, albeit limited effect of
grammatical errors on parsing. This outcome contrasts a 
study by Geertzen et al. \shortcite{geertzen2013} which reported a larger
performance gap of 7.6 UAS and 8.8 LAS between sentences with and without
grammatical errors. 
We believe that our analysis provides a more accurate estimate of this impact, 
as it controls for both sentence content and sentence length. The latter factor is crucial, 
since it correlates positively with the number of grammatical errors in the sentence, and negatively with 
parsing accuracy. 

\section{Related Work}
\label{related_work}

Previous studies on learner language proposed several annotation 
schemes for both POS tags and syntax \cite{hirschmann2007syntactic,diaz2010towards,ragheb2013,rosen2014evaluating}. 
The unifying theme in these proposals is a multi-layered analysis aiming 
to decouple the observed language usage from conventional structures in the 
foreign language. 

In the context of ESL, D{\i}az et al. \shortcite{diaz2010towards} propose
three parallel POS tag annotations for the \emph{lexical}, \emph{morphological} and \emph{distributional} 
forms of each word. In our work, we adopt the distinction between 
morphological word forms, which roughly correspond to our literal word readings, and 
distributional forms as the error corrected words. However, we account for morphological forms  
only when these constitute valid existing PTB POS tags and are contextually 
plausible. Furthermore, while the internal structure of invalid word forms 
is an interesting object of investigation, we believe that it is more suitable for 
annotation as word features rather than POS tags. Our treebank supports the addition 
of such features to the existing annotations.

The work of Ragheb and Dickinson \shortcite{dickinson2009dependency,ragheb2012defining,ragheb2013} 
proposes ESL annotation guidelines for POS tags and syntactic dependencies
based on the CHILDES annotation framework. This approach, called 
``morphosyntactic dependencies'' is related to our annotation scheme in its
focus on surface structures. Differently from this proposal, our annotations are grounded
in a parallel annotation of grammatical errors and include an additional layer of analysis
for the corrected forms. Moreover, we refrain from introducing new syntactic 
categories and dependency relations specific to ESL, thereby supporting computational 
treatment of ESL using existing resources for standard English. At the same time, 
we utilize a multilingual formalism which, in conjunction with our literal annotation 
strategy, facilitates linking the annotations to native language syntax.

While the above mentioned studies focus on annotation guidelines, attention
has also been drawn to the topic of parsing in the learner language domain. 
However, due to the shortage of syntactic resources for ESL, much of the work 
in this area resorted to using surrogates for learner data. For example, 
in Foster \shortcite{foster2007treebanks} and Foster et al. \shortcite{foster2008} parsing experiments are carried out on
synthetic learner-like data, that was created by automatic insertion of 
grammatical errors to well formed English text. In Cahill et al. 
\shortcite{cahill2014} a treebank of secondary level native students 
texts was used to approximate learner text in order to 
evaluate a parser that utilizes unlabeled learner data. 

Syntactic annotations for ESL were previously developed by Nagata et al. \shortcite{nagata2011}, 
who annotate an English learner corpus with POS tags and shallow syntactic parses. 
Our work departs from shallow syntax to full syntactic analysis, 
and provides annotations on a significantly larger scale. Furthermore, 
differently from this annotation effort, our treebank covers a wide range of learner native 
languages. An additional syntactic dataset for ESL, currently not available publicly,  
are 1,000 sentences from the EFCamDat dataset \cite{geertzen2013}, annotated 
with Stanford dependencies \cite{de2008stanford}. This dataset was used to 
measure the impact of grammatical errors on parsing by comparing performance on sentences with grammatical 
errors to error free sentences. The TLE enables a more direct way of estimating the 
magnitude of this performance gap by comparing performance on the same sentences
in their original and error corrected versions. Our comparison suggests that the effect 
of grammatical errors on parsing is smaller that the one reported in this study.

\section{Conclusion}
\label{conclusion}

We present the first large scale treebank of learner language, 
manually annotated and double-reviewed for POS tags and universal dependencies. 
The annotation is accompanied by a linguistically motivated framework for handling 
syntactic structures associated with grammatical errors. Finally, we benchmark 
automatic tagging and parsing on our corpus, and measure the effect of grammatical 
errors on tagging and parsing quality. The treebank will support empirical study 
of learner syntax in NLP, corpus linguistics and second language acquisition.

\section{Acknowledgements}
We thank Anna Korhonen for helpful discussions and
insightful comments on this paper. We also thank Dora Alexopoulou, 
Andrei Barbu, Markus Dickinson, Sue Felshin, Jeroen Geertzen, Yan Huang, 
Detmar Meurers, Sampo Pyysalo, Roi Reichart and the anonymous reviewers for 
valuable feedback on this work.
This material is based upon work supported by the Center for Brains, Minds, 
and Machines (CBMM), funded by NSF STC award CCF-1231216.

\bibliography{acl2016}
\bibliographystyle{acl2016}

\end{document}